\documentclass[conference]{IEEEtran}
\IEEEoverridecommandlockouts

\usepackage{cite}
\usepackage{amsmath,amssymb,amsfonts}
\usepackage{algorithmic}
\usepackage{graphicx}
\usepackage{textcomp}
\usepackage{xcolor}
\usepackage{color,soul}
\definecolor{cerulean}{rgb}{0.0,0.48,0.65}
\definecolor{green}{rgb}{0.01, 0.75, 0.24}
\definecolor{Black}{RGB}{0.0, 0.0, 0.0}
\newcommand{\red}[1]{\textcolor{red}{#1}}

\newcommand{\shadow}[1]{}

\def\r{\red}

\def\s{\shadow}

\definecolor{Boulder}{rgb}{0.6,0.6,0.6}
\definecolor{Silver}{rgb}{0.752,0.752,0.752}
\def\BibTeX{{\rm B\kern-.05em{\sc i\kern-.025em b}\kern-.08em
    T\kern-.1667em\lower.7ex\hbox{E}\kern-.125emX}}
\begin{document}

\title{Unveiling the Flaws: A Critical Analysis of Initialization Effect on Time Series \\ Anomaly Detection
}

\author{\IEEEauthorblockN{Alex Koran}
\IEEEauthorblockA{
\textit{McGill University}\\
\textit{MILA - Quebec AI Institute}\\
Montreal, QC, Canada \\
alexander.koran@mail.mcgill.ca}
\and
\IEEEauthorblockN{Hadi Hojjati}
\IEEEauthorblockA{\textit{McGill University}\\
\textit{MILA - Quebec AI Institute}\\
Montreal, QC, Canada \\
hadi.hojjati@mcgill.ca}
\and
\IEEEauthorblockN{Narges Armanfard}
\IEEEauthorblockA{\textit{McGill University}\\
\textit{MILA - Quebec AI Institute}\\
Montreal, QC, Canada \\
narges.armanfard@mcgill.ca}
}

\maketitle

\begin{abstract}
Deep learning for time-series anomaly detection (TSAD) has gained significant attention over the past decade. Despite the reported improvements in several papers, the practical application of these models remains limited. Recent studies have cast doubt on these models, attributing their results to flawed evaluation techniques. However, the impact of initialization has largely been overlooked. This paper provides a critical analysis of the initialization effects on TSAD model performance. Our extensive experiments reveal that TSAD models are highly sensitive to hyperparameters such as window size, seed number, and normalization. This sensitivity often leads to significant variability in performance, which can be exploited to artificially inflate the reported efficacy of these models. We demonstrate that even minor changes in initialization parameters can result in performance variations that overshadow the claimed improvements from novel model architectures. Our findings highlight the need for rigorous evaluation protocols and transparent reporting of preprocessing steps to ensure the reliability and fairness of anomaly detection methods. This paper calls for a more cautious interpretation of TSAD advancements and encourages the development of more robust and transparent evaluation practices to advance the field and its practical applications.
\end{abstract}

\begin{IEEEkeywords}
Time-Series Anomaly Detection, Anomaly Detection, Temporal Learning, Reliable Machine Learning.
\end{IEEEkeywords}

\section{Introduction}

Time series anomaly detection (TSAD) has been a significant area of interest in the field of data science due to its significant applications across various domains, including transportation \cite{ecml}, healthcare \cite{Ho_Armanfard_2023}, and industrial monitoring \cite{icassp}. The recent surge in interest is partly driven by the advancements in deep learning, which have shown promising results in handling time series data for tasks such as classification and forecasting \cite{tsforcast}.
In recent years, most research in time-series anomaly detection focused on modifying visual anomaly detection frameworks by incorporating a customized temporal architecture to handle sequential data. Early experiments have hinted that these algorithms can significantly outperform the state-of-the-art \cite{choi2021deep}. However, recent studies have cast a shadow of doubt over these claims \cite{garg}. They have shown that the performance of deep time-series models is not necessarily better than traditional time-series anomaly detection algorithms or even, in some cases, random baselines \cite{Kim_Choi_Choi_Lee_Yoon_2022}. The superior performance that some papers have reported can be attributed to other factors, such as incorrect evaluation procedures and inappropriate benchmark detests \cite{flawed}. As a result, most of the recent progress in deep time-series anomaly detection is practically useless in many real-life applications.
Initialization is one of the key aspects of time-series analysis that potentially plays an important role in the performance and reliability of TSAD algorithms. It can lead to misleading conclusions about the efficacy of TSAD algorithms, as the initialization steps can significantly affect the anomaly detection results. This phenomenon challenges the reliability of a TSAD method, as it makes it difficult to distinguish between genuine performance improvements due to the model and variations in results induced by initialization.
Despite the importance of initialization, its impact on anomaly detection performance has not been thoroughly investigated. In this paper, we provide a critical analysis of how initialization affects TSAD performance. We aim to uncover the flaws and biases introduced by commonly employed initialization and preprocessing techniques. Through extensive experiments, we demonstrate how different initialization techniques can significantly alter detection outcomes, highlighting the risk of artificial performance boosts resulting from variations in initialization.
Our findings suggest that while initialization is an integral part of time series processing, it must be applied judiciously, with a clear understanding of its potential impact on anomaly detection results. By shedding light on the complexities and pitfalls of initialization in time series anomaly detection, we hope to contribute to the development of more robust and reliable detection methodologies, ultimately advancing the field and its practical applications.

\section{Experiment Setup}

To demonstrate the effect of various initialization techniques on the performance of TSAD methods, we designed a comprehensive experimental setup which involved systematically varying key initialization parameters on benchmark datasets.

\subsection{Datasets}

\subsubsection{Secure Water Treatment (SWaT) Dataset \cite{Goh2016}}
The SWaT dataset is derived from a water treatment testbed built at the Singapore University of Technology and Design. It is designed to simulate a real-world water treatment process and contains data that represent normal operations as well as various attack scenarios that cause anomalies. The dataset includes a comprehensive range of sensor readings and actuator statuses collected over a period of days. SWAT is considered to be a standard benchmark, yet similar to other TSAD datasets, SWAT suffers from issues such as labelling discrepancy. \s{The data is labeled, distinguishing between normal and anomalous states, making it ideal for evaluating anomaly detection algorithms. \r{DO we want to provide a quick discussion that it is not the best dataset!! what is the main issue with SWAT? Feels awkward to consider SWAT. }}

\subsubsection{Server Machine Dataset (SMD) \cite{Su2019}}
The SMD dataset consists of multivariate time-series data collected from 28 server machines in a large internet company over five weeks. Each server is monitored by 33 sensors, capturing metrics such as CPU load, network usage, and memory usage. This dataset is particularly valuable due to its scale and the diversity of operational conditions it encompasses. Following \cite{MOSAD}, in our study, we modified the dataset and used an interval-segmented version of SMD. This approach increases the complexity of the dataset and helps mitigate the issue of triviality.

\s{REDUNDANT. REMOVE. Both datasets provide a \r{robust} framework for testing and validating the effectiveness of time-series anomaly detection methods. The SWaT dataset focuses on a physical industrial control system, providing insights into the detection of anomalies in critical infrastructure. In contrast, the SMD dataset offers a view into the operational metrics of server machines, allowing for the evaluation of methods in an IT context. Together, these datasets help ensure that the tested methods can generalize across different types of time-series data and anomaly scenarios.}

\subsection{Methods}

For our experiment, we used three state-of-the-art (SOTA) methods commonly employed by TSAD researchers for comparison and benchmarking over the past few years.

\subsubsection{Graph Deviation Network (GDN)}

GDN \cite{deng2021graph} is designed to detect anomalies in multivariate time series data by leveraging the inherent structure of the data. The idea behind GDN is to model the relationships between different time series (sensors) using a graph structure where each node represents a time series, and edges capture the dependencies between them. GDN uses Graph Neural Networks (GNNs) to learn these dependencies and identify deviations from normal behavior.
The network is trained to predict the value of each sensor at a given time step based on the values of the other sensors and itself in the past. Anomalies are detected when the actual sensor values significantly deviate from the predicted values. This method allows GDN to capture and leverage the spatial dependencies in multivariate time series data.

\subsubsection{Multivariate Time-Series Anomaly Detection via Graph Attention Networks (MTAD-GAT)}

MTAD-GAT \cite{MTADGAT} employs a combination of convolutional layers, graph attention networks, and recurrent units to capture both spatial and temporal dependencies in multivariate time series data. The architecture begins with a 1-D convolutional layer\s{         GAT IS using some sort of feature extraction (using 1-D CNN) while GDN is working with raw data. A discussion on this aspect could be interesting      e.g. A more stable trend of GAT could be associated with this matter???} that extracts high-level features from each time series, enabling the network to understand the inherent patterns within the data. Following this, two parallel Graph Attention Networks (GAT) layers are utilized: one focusing on feature-oriented dependencies and the other on time-oriented dependencies. This dual approach enables the model to capture the intricate relationships between different features and across different time steps. The outputs from the convolutional and GAT layers are then fed into a Gated Recurrent Unit (GRU), which models the sequential patterns in the data, providing a deeper understanding of temporal dependencies. 

\subsubsection{Unsupervised Anomaly Detection (USAD)}

USAD \cite{audibert2020usad} is an unsupervised anomaly detection method for multivariate time series, utilizing a combination of autoencoders and adversarial training. Initially, the method involves training an autoencoder to minimize reconstruction errors, which helps learn the normal patterns present in the data. Following this, USAD incorporates adversarial training, where the objective is to differentiate between the actual time series data and the reconstructed data generated by the autoencoder. This adversarial phase aims to amplify the reconstruction error when anomalies are present, thus making the model more sensitive to deviations.

\section{Results and Discussion}

\subsection{Window Size Effect}

Window size is a critical parameter in TSAD as it defines the context for anomaly detection. Choosing an appropriate window size is essential because it influences the model's ability to capture temporal dependencies. 

\begin{figure}
\vspace{-1em}
\includegraphics[width=0.48\textwidth]{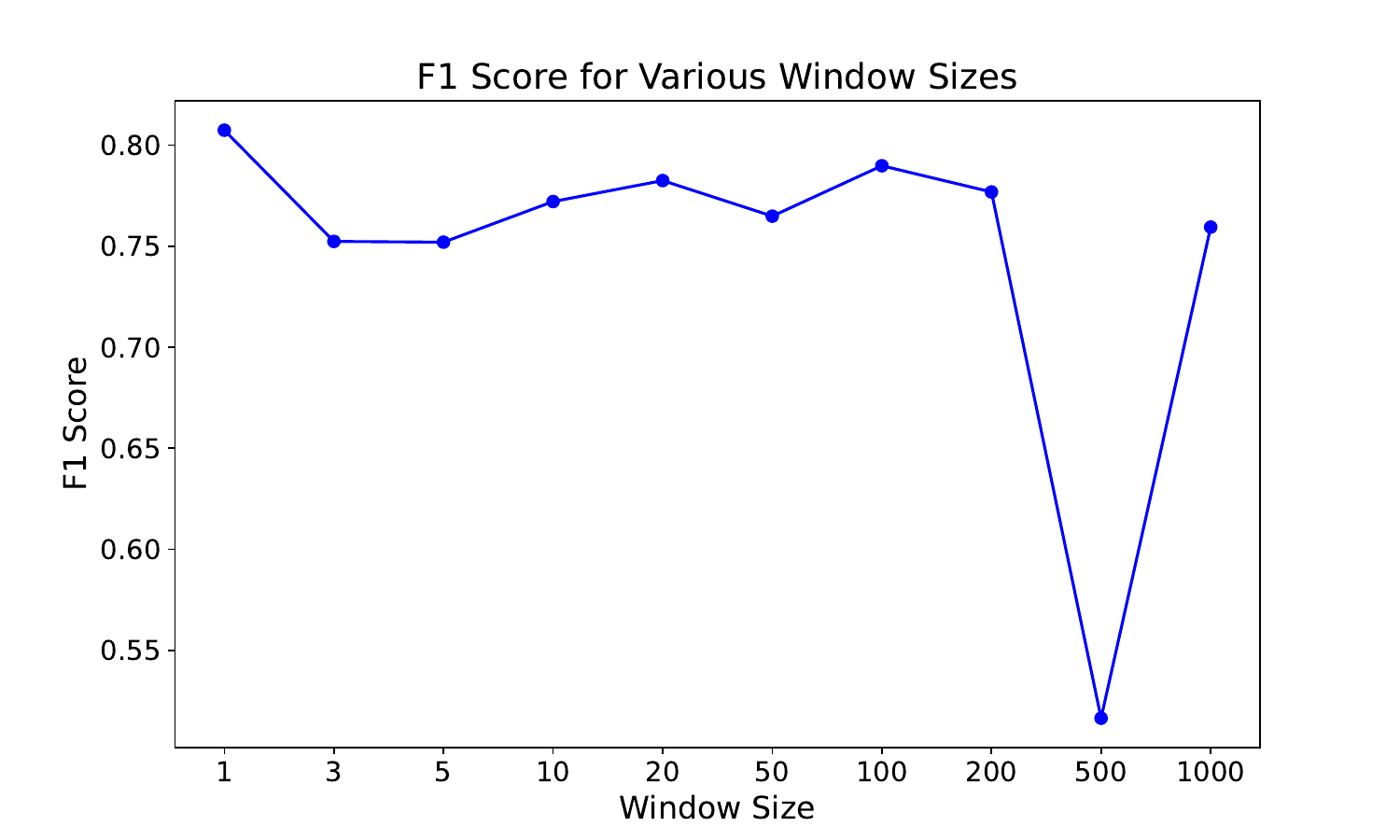}
\vspace{-1em}
\caption{F1 scores obtained by GDN on the SWAT dataset for varying window sizes.} \label{gdn_swat_window}
\end{figure}

A small window size captures fine-grained temporal patterns but may miss broader trends and context. Conversely, a larger window size captures broader temporal trends but may smooth over important short-term anomalies. The choice of window size should balance the need to capture relevant patterns while avoiding the dilution of critical anomalies. Ideally, a reliable method should not be hypersensitive to this parameter: very minor adjustments in window size should not significantly impact performance in an ideal scenario. 
To assess the effect of window size on the anomaly detection performance of our TSAD methods, we set it to 1, 3, 5, 10, 20, 50, 100, 200, 500 and 1000 to represent a wide range of possible values. The resulting figure is plotted in Figure \ref{gdn_swat_window}.


In this figure, we observe the F1 scores for various window sizes when running GDN on the SWAT dataset. The variability in the model's performance across different window sizes is immediately noticeable. For instance, changing the window size from 1 to 3 results in more than a $5\%$ decrease in F1 score. This should not occur; in fact, we should expect the model to perform better when provided with more data within the window. It is also worth noting that even a $3\%$ improvement is considered groundbreaking in the literature.
The sharp drop in performance for a window size of $500$ is also alarming. Although several trials suggest it may be an outlier, this is still far from ideal, as an anomaly detection model in practice should not have drastic dips in performance merely due to changing a parameter.

Not only is there a lack of correlation between window size and performance, but using a window size of 1 actually yielded the highest F1 score in this experiment. This is unexpected because GDN is a predictive method, and a window size of 1 means the model only analyzes the previous time point when determining if a given time point is anomalous.
This suggests that GDN on SWAT is not effectively capturing long-term dependencies when using larger window sizes; if it were, we would expect better performance for window sizes above 1, at least up to a certain point.
We performed similar window size experiments on the MTAD-GAT and USAD models and observed the same trend.
Our main takeaway is that these models do not sufficiently utilize their window size architecture to capture long-term dependencies in datasets such as SWAT. Additionally, they suffer from large performance variance relative to the window size, often much more than 5\%. This variation in performance is far greater than what current methods achieve over traditional methods, challenging the reliability of their claims, as the observed performance boost might be attributed to changing a parameter such as window size. 
It is also worth noting that there is no consistency in the literature regarding the standard window size for a dataset. Furthermore, some papers claim to downsample the data due to memory concerns, which also impacts the amount of information in each window and can drastically affect performance compared to the existing SOTA methods.

\subsection{Seed Variability}

Many model initialization and data preprocessing steps involve randomness. To ensure the reproducibility of results, it is crucial to evaluate the impact of seed variability. Running experiments with different seeds helps in understanding the stability of the initialization methods and the algorithms' performance. Ideally, changing the seed number should not significantly affect performance. If it does, it suggests that the method is not reliable and that optimal seed selection might have been done using grid search to inflate the model's performance compared to established baselines.
In our experiments, we ran the models with their default parameters but fixed the seed to various seed numbers, namely $s \in \{1, 2, \ldots, 25\}$, although any $25$ different seeds would work. The goal was to examine the variability of the models.\s{, some of which was not apparent when running seed variability experiments on only $10$ different seeds.}



\begin{figure}
\vspace{-1em}
\includegraphics[width=0.48\textwidth]{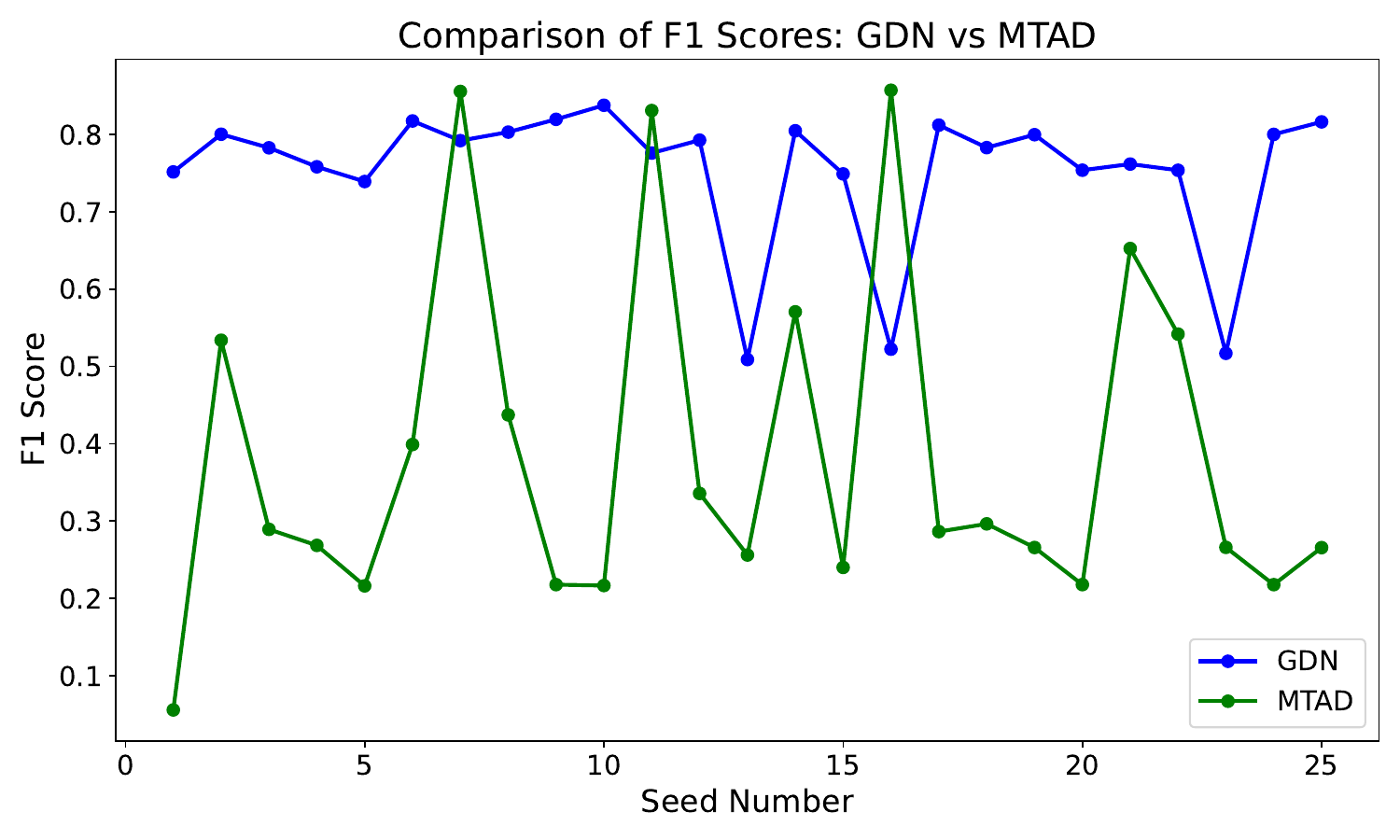}
\vspace{-1em}
\caption{F1 scores obtained by GDN and MTAD-GAT on the SWAT dataset for varying random seed numbers.} \label{gdn_mtad_swat_seed}
\end{figure}

Figure \ref{gdn_mtad_swat_seed} illustrates 25 trials of GDN on SWAT with the exact same initial parameters. Among the first 10 seeds, there are sharp increases and decreases, many exceeding $5\%$ changes or more. This already supports our claim from the previous section that the model's results are too variable and the models are hypersensitive to parameters, which should not be the case.
Upon inspecting and plotting $25$ trials, we observe even more drastic changes in performance. In 3 of these 25 seeds, the model's performance dropped below a 0.55 F1 score, representing more than a $25\%$ change from the other F1 values. This is very alarming, as we would expect a reliable model to be robust across multiple trials with the same input parameters.
%
Figure \ref{gdn_mtad_swat_seed} also displays the same trend but with the MTAD-GAT model:\s{In these experiments the anomaly threshold is set to the maximum anomaly score in the validation set, similar to the other experiments and to GDN's implementation.} Some trials yield F1 scores above $0.8$ while others yield F1 scores below $0.3$. We observe once again that with no input parameter change, there is a discrepancy between the F1 and overall performance of MTAD-GAT across different trials.
The results of these experiments reiterate the point raised in the previous section: current SOTA models seem to be overly sensitive to hyperparameters such as seed number or window size. This sensitivity can provide a shortcut to improved performance simply by searching for and reporting the best parameters for the model rather than for the baselines. We observe performance changes as high as double digits just by perturbing the window size or changing the seed number, which is far greater than the claimed improvements due to their innovative architecture compared to SOTA. This casts a shadow of doubt over which factor plays the major role in the observed improvements: the model itself or the initialization parameters.






\subsection{Normalization} 

Normalization is the process of scaling data to a fixed range, typically between 0 and 1, or to have a mean of 0 and a standard deviation of 1. This step is crucial for ensuring that different features of time series data contribute equally to the anomaly detection process. Without normalization, features with larger numerical ranges can disproportionately influence the model's performance, leading to biased results. Proper normalization ensures that all features contribute equally to the anomaly detection task, improving the model's accuracy and efficiency. However, inappropriate normalization can smooth out anomalies or fail to capture critical variations in the data, leading to decreased detection performance.

\begin{figure}
\vspace{-1em}
\includegraphics[width=0.48\textwidth]{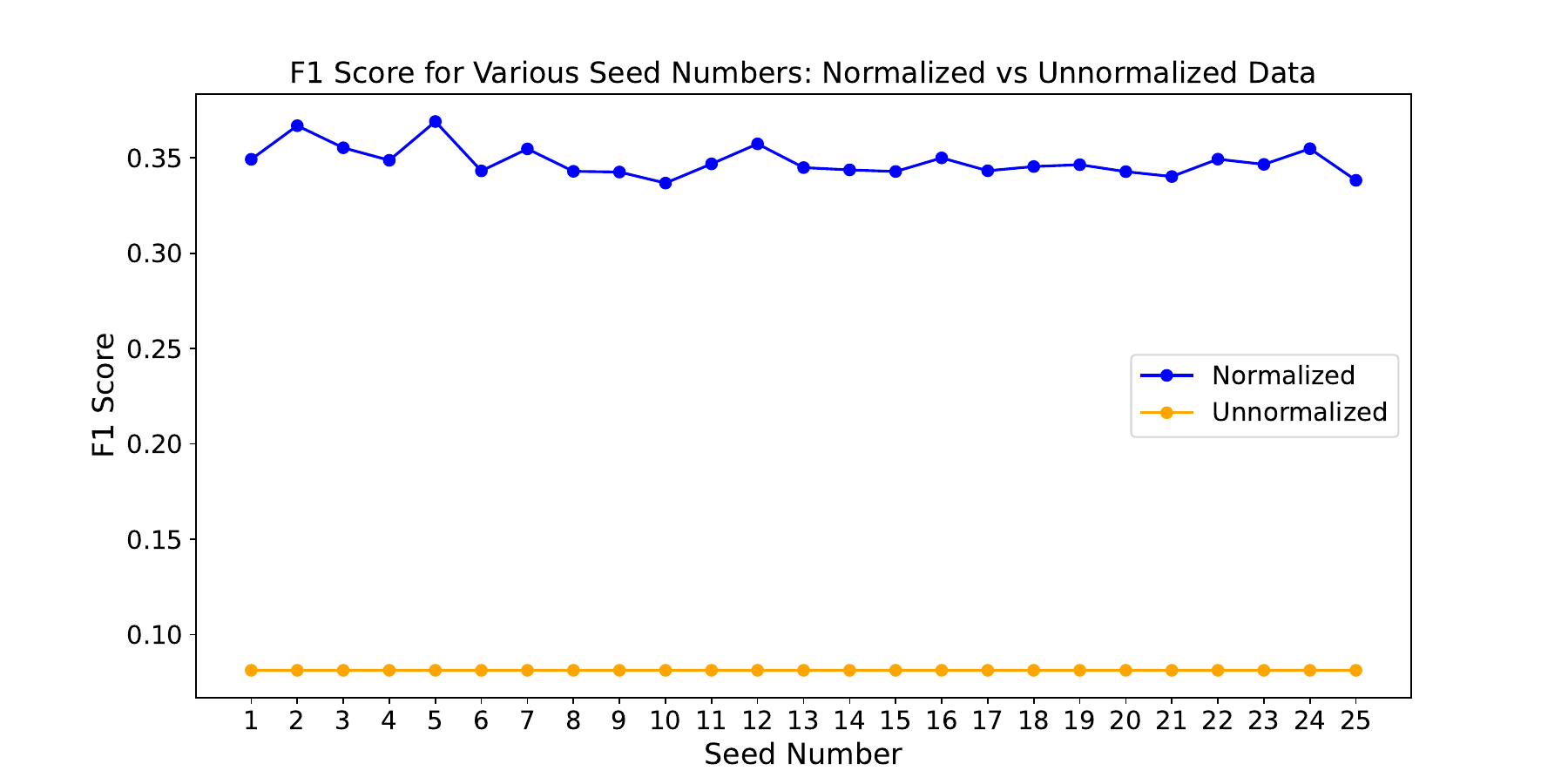}
\vspace{-1em}
\caption{F1 scores obtained by USAD on the SWAT dataset for varying random seed numbers. The blue line is the USAD implementation which has normalization, whereas the orange line is USAD without using normalization.} \label{usad_seed_normalized_vs_unnormalized}
\end{figure}

In our experiments, we ran USAD on the SWAT dataset, both with and without normalization, across 25 different seeds. Figure \ref{usad_seed_normalized_vs_unnormalized} illustrates the F1 scores obtained from these trials. The results show that normalization significantly impacts the performance and variability of the USAD model. With normalization, the model achieved far higher F1 scores, indicating vast improvement in performance. However, the model also exhibited greater variability, suggesting that normalization might introduce sensitivity to the initial parameters.

These results also hint that the model is extremely sensitive to normalization. This sensitivity can be exploited to artificially inflate the performance of an TSAD method. By selecting specific normalization techniques or tuning parameters to favor the model, one can achieve seemingly superior results that may not generalize well to other datasets or real-world applications. The inconsistency in performance due to normalization underscores the need for rigorous evaluation and transparent reporting of the input data's characteristics and preprocessing steps to ensure the reliability and fairness of anomaly detection methods.






\subsection{Analysis of Anomaly Score}

As observed in previous sections, many seemingly unrelated parameters can significantly affect the performance of a TSAD model. As several studies have hinted earlier, the current numerical improvements many papers reported are often due to flawed thresholding and evaluation protocols. To have a more realistic view of the true output of TSAD models under different conditions, we analyze the density of raw anomaly scores before applying any transformation, thresholding, or point adjustment. We plot the density for training, validation, and test sets: in the SWAT and SMD datasets, the training and validation sets consist only of normal data, while the test set includes both normal and abnormal data. Ideally, we would like to observe significantly low anomaly scores for the training, validation, and normal test data, and significantly higher anomaly scores for the abnormal test data.

\begin{figure}
\vspace{-1em}
\includegraphics[width=0.48\textwidth]{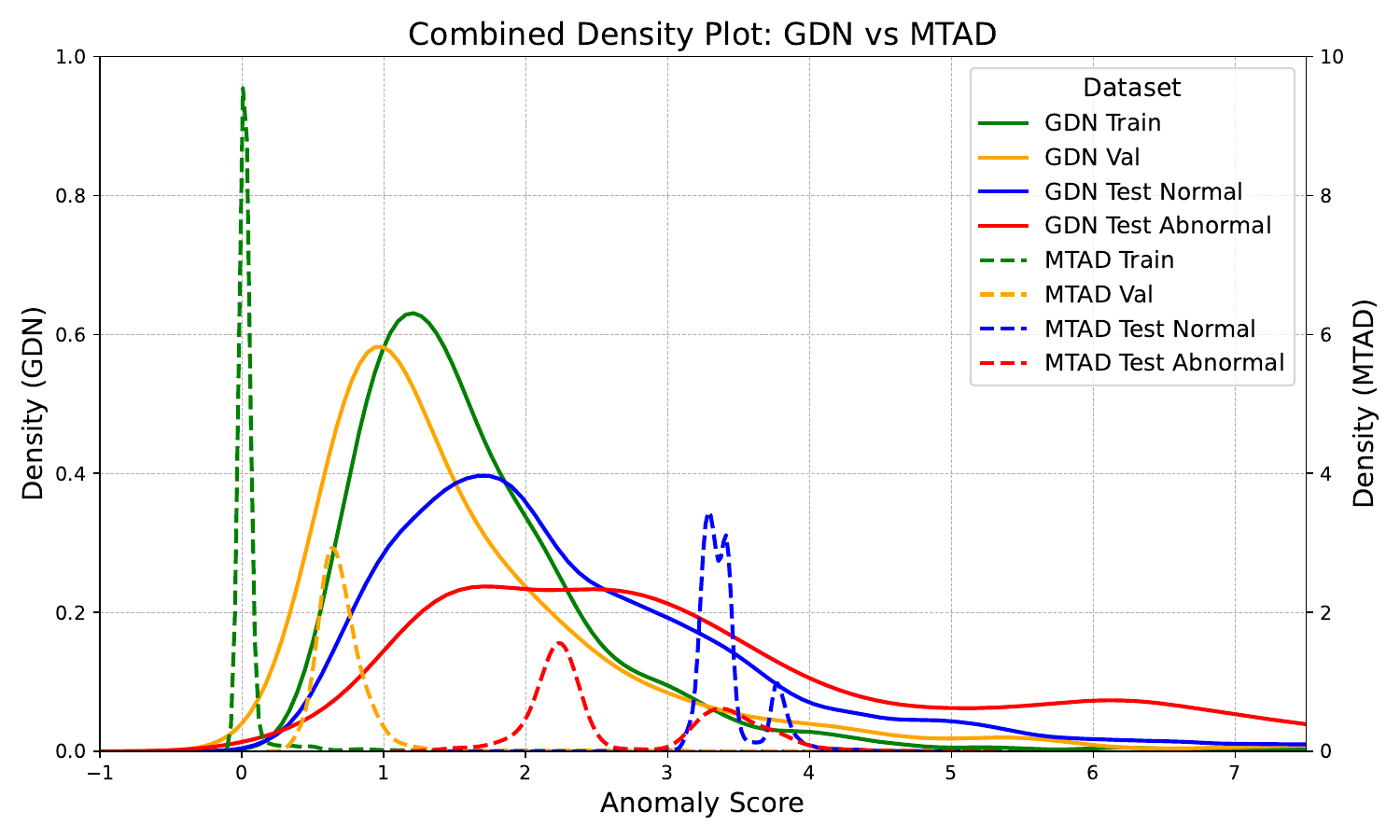}
\vspace{-1em}
\caption{Density plots of anomaly scores obtained by GDN on the SMD dataset and MTAD on the SWAT dataset. The plot illustrates the distribution of anomaly scores across different sets: the training set, validation set, normal data from the test set, and abnormal data from the test set.} \label{gdn_mtad_density}
\end{figure}

Figure \ref{gdn_mtad_density} illustrates the anomaly score density of GDN on the SMD dataset. We can observe that the anomaly scores of the training samples are what we expect: near zero, since the model learned during training that these were normal. The validation set, which consists of unseen normal data, also shows anomaly scores close to zero.
However, the density of the anomaly scores for abnormal samples during the test phase does not significantly differ from the normal data, and is still close to zero. This explains why the model does not perform well in this experimental setup: it cannot distinguish between normal and abnormal samples. Yet, by incorporating schemes such as point adjustment, it might be possible to report an acceptable F1 score.
Another interesting and extreme example of counterintuitive anomaly score densities is observed with MTAD-GAT on the SWAT dataset. As demonstrated by the dashed lines in Figure \ref{gdn_mtad_density}, the anomaly scores of the normal test data are even larger than those of the abnormal data. This is counterintuitive given that the model is trained to minimize the anomaly score on normal data. However, since the density of normal test data scores does not fully overlap with the distribution of abnormal anomaly scores, the model could still achieve a good F1 score by using a kernel to transform the anomaly scores and thresholding to isolate the densities. This highlights how post-processing techniques can mask underlying model deficiencies and produce seemingly favorable results.
At the cost of sacrificing the robustness of the model, one could thoroughly search for a set of optimal initialization parameters such as window size, seed number, and normalization technique. Combined with existing flawed evaluation protocols, this approach could potentially outperform any SOTA method without actually improving the model's impact on real-world TSAD problems. This emphasizes the need for rigorous evaluation and transparent reporting of preprocessing steps and input data characteristics to ensure the reliability and fairness of anomaly detection methods.




\section{Conclusion}

This paper critically analyzed the impact of initialization on TSAD models. Our experiments revealed that they are highly sensitive to parameters like window size, seed number, and normalization, leading to significant performance variability. This sensitivity can be exploited to artificially inflate performance results, raising concerns about the robustness and reliability of current TSAD methods. Our findings highlight the necessity for rigorous evaluation protocols and transparent reporting of preprocessing steps to ensure the reliability and fairness of anomaly detection methods.
To advance the field, we recommend that future research focuses on developing more robust models that are less sensitive to initialization parameters. Additionally, standardizing evaluation protocols and ensuring transparency in reporting preprocessing steps are essential to prevent the exploitation of these sensitivities.

\pagebreak
\fontsize{16pt}{18pt}\selectfont
\bibliographystyle{IEEEbib}
\bibliography{refs}

\begin{thebibliography}{10}

\bibitem{ecml}
Hadi Hojjati, Mohammadreza Sadeghi, and Narges Armanfard,
\newblock ``Multivariate time-series anomaly detection with temporal self-supervision and graphs: Application to vehicle failure prediction,''
\newblock in {\em Machine Learning and Knowledge Discovery in Databases: Applied Data Science and Demo Track}, Gianmarco De~Francisci~Morales, Claudia Perlich, Natali Ruchansky, Nicolas Kourtellis, Elena Baralis, and Francesco Bonchi, Eds., Cham, 2023, pp. 242--259, Springer Nature Switzerland.

\bibitem{Ho_Armanfard_2023}
Thi Kieu~Khanh Ho and Narges Armanfard,
\newblock ``Self-supervised learning for anomalous channel detection in eeg graphs: Application to seizure analysis,''
\newblock {\em Proceedings of the AAAI Conference on Artificial Intelligence}, vol. 37, no. 7, pp. 7866--7874, Jun. 2023.

\bibitem{icassp}
Hadi Hojjati and Narges Armanfard,
\newblock ``Self-supervised acoustic anomaly detection via contrastive learning,''
\newblock in {\em ICASSP 2022 - 2022 IEEE International Conference on Acoustics, Speech and Signal Processing (ICASSP)}, 2022, pp. 3253--3257.

\bibitem{tsforcast}
Jos\'{e}~F. Torres, Dalil Hadjout, Abderrazak Sebaa, Francisco Mart\'{\i}nez-\'{A}lvarez, and Alicia Troncoso,
\newblock ``Deep learning for time series forecasting: A survey,''
\newblock {\em Big Data}, vol. 9, no. 1, pp. 3--21, 2021,
\newblock PMID: 33275484.

\bibitem{choi2021deep}
Kukjin Choi, Jihun Yi, Changhwa Park, and Sungroh Yoon,
\newblock ``Deep learning for anomaly detection in time-series data: Review, analysis, and guidelines,''
\newblock {\em IEEE access}, vol. 9, pp. 120043--120065, 2021.

\bibitem{garg}
Astha Garg, Wenyu Zhang, Jules Samaran, Ramasamy Savitha, and Chuan-Sheng Foo,
\newblock ``An evaluation of anomaly detection and diagnosis in multivariate time series,''
\newblock {\em IEEE Transactions on Neural Networks and Learning Systems}, vol. 33, no. 6, pp. 2508--2517, 2022.

\bibitem{Kim_Choi_Choi_Lee_Yoon_2022}
Siwon Kim, Kukjin Choi, Hyun-Soo Choi, Byunghan Lee, and Sungroh Yoon,
\newblock ``Towards a rigorous evaluation of time-series anomaly detection,''
\newblock {\em Proceedings of the AAAI Conference on Artificial Intelligence}, vol. 36, no. 7, pp. 7194--7201, Jun. 2022.

\bibitem{flawed}
Renjie Wu and Eamonn~J. Keogh,
\newblock ``Current time series anomaly detection benchmarks are flawed and are creating the illusion of progress,''
\newblock {\em IEEE Transactions on Knowledge and Data Engineering}, vol. 35, no. 3, pp. 2421--2429, 2023.

\bibitem{Goh2016}
J.~Goh, S.~Adepu, K.~N. Junejo, and A.~Mathur,
\newblock ``A dataset to support research in the design of secure water treatment systems,''
\newblock in {\em The 11th International Conference on Critical Information Infrastructures Security}, 2016.

\bibitem{Su2019}
Y.~Su, Y.~Zhao, C.~Niu, R.~Liu, W.~Sun, and D.~Pei,
\newblock ``Robust anomaly detection for multivariate time series through stochastic recurrent neural network,''
\newblock in {\em Proceedings of the 25th ACM SIGKDD International Conference on Knowledge Discovery \& Data Mining}, A.~Teredesai, V.~Kumar, Y.~Li, R.~Rosales, E.~Terzi, and G.~Karypis, Eds., 2019, pp. 2828--2837.

\bibitem{MOSAD}
T.~Lai, T.~K.~K. Ho, and N.~Armanfard,
\newblock ``Open-set multivariate time-series anomaly detection,''
\newblock in {\em Proceedings of the 27th European Conference on Artificial Intelligence}, 2024.

\bibitem{deng2021graph}
Ailin Deng and Bryan Hooi,
\newblock ``Graph neural network-based anomaly detection in multivariate time series,''
\newblock in {\em Proceedings of the AAAI Conference on Artificial Intelligence}, 2021, vol.~35, pp. 4027--4035.

\bibitem{MTADGAT}
H.~Zhao, Y.~Wang, J.~Duan, C.~Huang, D.~Cao, Y.~Tong, B.~Xu, J.~Bai, J.~Tong, and Q.~Zhang,
\newblock ``Multivariate time-series anomaly detection via graph attention network,''
\newblock in {\em 2020 IEEE International Conference on Data Mining (ICDM)}, Los Alamitos, CA, USA, nov 2020, pp. 841--850, IEEE Computer Society.

\bibitem{audibert2020usad}
Julien Audibert, Pietro Michiardi, Francis Guyard, Stephan Marti, and Marc~A. Zuluaga,
\newblock ``Usad: Unsupervised anomaly detection on multivariate time series,''
\newblock in {\em Proceedings of the 26th ACM SIGKDD International Conference on Knowledge Discovery \& Data Mining}, Virtual Event, CA, USA, August 23-27 2020, ACM, pp. 3395--3404.

\end{thebibliography}

\end{document}